\documentclass{article}
\usepackage{spconf,amsmath,epsfig}
\usepackage{url}

\usepackage[acronym]{glossaries}
\newacronym{eo}{EO}{Earth Observation}
\newacronym{utm}{UTM}{Universal Transverse Mercator}

\title{Major TOM: Expandable Datasets for Earth Observation}
\name{Alistair Francis, Mikolaj Czerkawski}

\address{$\Phi$-lab\\
	European Space Agency}
\usepackage{lipsum}
\begin{document}
%
\maketitle
\begin{abstract}
    Deep learning models are increasingly data-hungry, requiring significant resources to collect and compile the datasets needed to train them, with Earth Observation (EO) models being no exception. However, the landscape of datasets in EO is relatively atomised, with interoperability made difficult by diverse formats and data structures. If ever larger datasets are to be built, and duplication of effort minimised, then a shared framework that allows users to combine and access multiple datasets is needed. Here, Major~TOM (Terrestrial Observation Metaset) is proposed as this extensible framework. Primarily, it consists of a geographical indexing system based on a set of grid points and a metadata structure that allows multiple datasets with different sources to be merged. Besides the specification of Major~TOM as a framework, this work also presents a large, open-access dataset, MajorTOM-Core, which covers the vast majority of the Earth's land surface. This dataset provides the community with both an immediately useful resource, as well as acting as a template for future additions to the Major~TOM ecosystem.\\\\\textit{Access:}\\\url{https://huggingface.co/Major-TOM}
\end{abstract}

    \begin{figure*}[t]
        \centering
        \includegraphics[width=\textwidth]{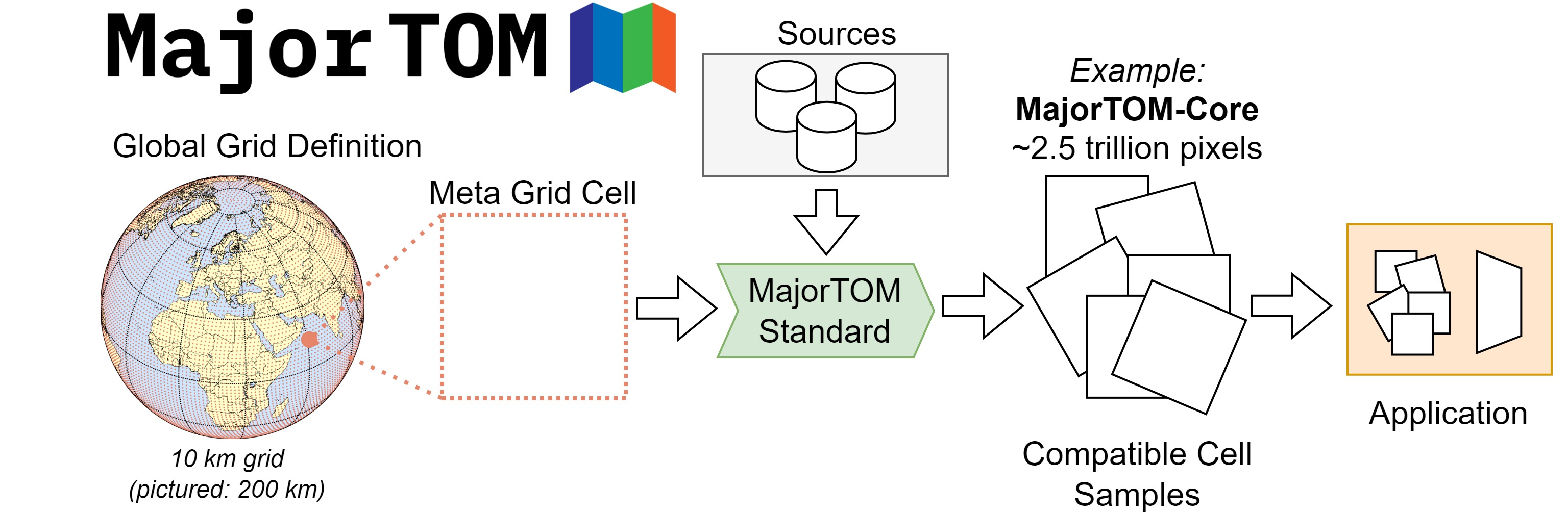}
        \caption{Major TOM paves the way for a standardized definition of AI-oriented datasets by relying on a specific grid standard. As an example, the MajorTOM-Core dataset delivers over 2 trillion pixels of Sentinel-2 data in total, spanning across nearly every piece of land captured by Sentinel-2.}
        \label{fig:intro-figure}
    \end{figure*}

\section{Introduction}
\label{sec:intro}
    \gls{eo} stands out as a field of knowledge with some of the largest sources of visual data ever produced. For example, the archive of the Copernicus Sentinel data was 40~PB at the end of 2022~\cite{CopernicusData}. Whilst these archives continue to grow at pace, and despite the widely accepted insight that data is one of the most influential factors affecting the performance of deep learning models, the size of the datasets that most models are trained on is often severely limited when compared to other fields of computer vision. For example, datasets like LAION-5B~\cite{laion_dataset,LAION_EO} containing 5 billion image-text pairs are used to train massive models (such as StableDiffusion~\cite{Rombach_2022_CVPR}). There is a growing understanding that the scaling of large \gls{eo} models will require techniques that leverage the underlying structures and contextual information that \gls{eo} data has to formulate successful pretraining tasks (e.g. geographically aware learning~\cite{satclip}, ground image alignment~\cite{graft_dataset}, or multitemporal masking and reconstruction~\cite{satmae}). 

    In recent years, the means by which users interact with \gls{eo} data archives have been evolving rapidly. Platforms and services now provide API access to data, at varying levels of speed, coverage, and price, such as the Copernicus Data Space, CREODIAS, SentinelHub, Google Earth Engine, and the Microsoft Planetary Computer. Gone are the days of selecting individual scenes and downloading them in their entirety, as users are now able to download specific bands across specific geographical extents that can be much smaller (and thus faster) than the entire product. However, few (if any) models have been trained directly on data \textit{streamed} from these archives. Instead, recent efforts to train large-scale \gls{eo} use these services to download massive quantities of data, often converting it into a more AI-ready format (e.g. resampling to different resolutions, breaking into patches, etc.), before training a model on that static dataset.

    It is essential that these large datasets are made openly available. In this way, the time and effort expended on data collection can be additive, rather than competitive. Promisingly, many large datasets have been made available (see Sec.~\ref{sec:rel_work} for more details). At present, however, the community continues to create bespoke datasets for each project, which not only takes a significant amount of resources, but also leads to difficulties in comparing different approaches' performance and capabilities, or rapidly prototyping the same approach with many kinds of input data.

    Major TOM (Terrestial Observation Metaset) aims to address this, by aiding in the dissemination and utilisation of existing and new datasets through a shared, flexible format. With the combination of a simple geographical indexing grid, and some best practices regarding the way data is processed, stored and described, Major TOM is envisioned as a vehicle through which a diverse family of datasets can be created and used, either in isolation, or in combination, with flexibility (see Fig.~\ref{fig:intro-figure}). More information about the technical aspects of the framework can be found in Sec.~\ref{sec:design}. As well as the framework, this work also presents the largest openly available dataset of Sentinel-2 imagery created to date, called MajorTOM-Core, and the ongoing expansions to other modalities, which is described further in Sec.~\ref{sec:data}.

\section{Related Work}
\label{sec:rel_work}

    Since the training of large neural networks entered the mainstream (from 2012 onwards~\cite{alexnet}), many \gls{eo} datasets have been curated to support deep learning pipelines. However, the existing solutions do not achieve the size and adaptability of Major TOM. Apart from a few exceptions, this is because most datasets have been created with a specific narrow task in mind (such as scene classification, or land use monitoring) and not as part of a wider strategy to empower data-hungry deep learning solutions. For a more complete reference, the reader is referred to the recent surveys on this topic~\cite{moredata_schmitt,earthnets}.
    
    \begin{table}[]
        \centering
        \caption{Comparison of selected large \gls{eo} datasets. The number of pixels are used to indicate dataset volume.}
         \resizebox{\columnwidth}{!}{
        \begin{tabular}{llcrcr}
            \hline
            Dataset & Modality & Patch & Count ($\approx$) & Coverage & Gigapixels\\
            \hline
            BigEarthNet~\cite{bigearthnet_dataset} & \begin{tabular}{@{}l@{}}S2 L2A\\S1 GRD\end{tabular} & 120$\times$120 & 590,326 & Europe & 8.5\\
            \hline
            SEN12MS-CR~\cite{sen12mscr_dataset} & \begin{tabular}{@{}l@{}}S2 L1C\\S1 GRD\end{tabular} & 256$\times$256 & 122,218 & Global & 8.0\\
            \hline
            SEN12MS-CR-TS~\cite{sen12mscrts_dataset} & \begin{tabular}{@{}l@{}}S2 L1C\\S1 GRD\end{tabular} & 256$\times$256 & 467,340 & Global & 30.6\\
            \hline
            CloudSEN12~\cite{cloudsen12_dataset} & \begin{tabular}{@{}l@{}} S2 L1C \\ S2 L2A \\ S1 GRD\end{tabular} & 509$\times$509 & 49,400 & Global & 12.8\\
            \hline
               &   SPOT 6/7 & 1054$\times$1054 & & & 69.8 \\
            \begin{tabular}{@{}l@{}}WorldStrat~\cite{worldstrat_dataset}\\ ~ \end{tabular} & \begin{tabular}{@{}l@{}}S2 L1C\\S2 L2A\end{tabular} & 158$\times$158 & \begin{tabular}{@{}l@{}}62,832\\ ~ \end{tabular} &  \begin{tabular}{@{}l@{}}Global\\ ~ \end{tabular} & 25.1\\
            \hline
            SeCo~\cite{seco_dataset} & S2 L2A & 265$\times$265 & 1,000,000 & Global & 70.2\\
            \hline
            SSL4EO-S12~\cite{ssl4eo_dataset} & \begin{tabular}{@{}l@{}} S2 L1C \\ S2 L2A \\ S1 GRD\end{tabular} & 264$\times$264 & 1,000,000 & Global & 69.7\\
            \hline
            RapidAI4EO~\cite{rapidai4eo_dataset} & \begin{tabular}{@{}l@{}} S2 L2A \\ Planet \end{tabular} & \begin{tabular}{@{}c@{}}60$\times$60\\200$\times$200\end{tabular} & 500,000 & Europe & \begin{tabular}{@{}r@{}}21.6\\1460.0\end{tabular} \\
            \hline
            SATLAS~\cite{satlas_dataset} & \begin{tabular}{@{}l@{}}NAIP\\S2 L1C\end{tabular} & \begin{tabular}{@{}c@{}}8192$\times$8192\\512$\times$512\end{tabular} &  \begin{tabular}{@{}c@{}}46,000 \\856,000\end{tabular} & \begin{tabular}{@{}c@{}}USA\\Global\end{tabular}& \begin{tabular}{@{}r@{}}3087.0 \\224.4\end{tabular}\\
            \hline
            GRAFT~\cite{graft_dataset} & \begin{tabular}{@{}l@{}}NAIP\\S2 (RGB)\end{tabular} & 448$\times$448 & \begin{tabular}{@{}r@{}}10,200,000\\8,700,000\end{tabular} & \begin{tabular}{@{}c@{}}USA\\Global\end{tabular} & \begin{tabular}{@{}r@{}}2047.2\\1746.1\end{tabular}\\
            \hline
            \textbf{MajorTOM-Core} & \begin{tabular}{@{}l@{}}S2 L1C \\S2 L2A\\S1 RTC\end{tabular} & 1,068$\times$1,068 & \begin{tabular}{@{}l@{}}2,245,886 \\2,245,886\\1,469,955\end{tabular}& Global & \begin{tabular}{@{}l@{}}2561.7\\2561.7\\1676.7\end{tabular}\\
            \hline
        \end{tabular}
    }
        \label{tab:stats}
    \end{table}

    Tab.~\ref{tab:stats} features a selection of the existing large-scale datasets. In many cases, the datasets are either geographically constrained, such as BigEarthNet~\cite{bigearthnet_dataset} and RapidAI4EO~\cite{rapidai4eo_dataset} focusing on Europe, or, where distributed globally, sparsely sample the globe. In other cases, the number of samples combined with the patch size, quantified with the total number of pixels in the dataset, is limited. Usually, the pixel count is on the order of billions~\cite{bigearthnet_dataset,sen12mscr_dataset}, tens of billions~\cite{sen12mscrts_dataset,worldstrat_dataset,seco_dataset,rapidai4eo_dataset, ssl4eo_dataset, cloudsen12_dataset}, or hundreds of billions~\cite{satlas_dataset} (for the Sentinel-2 sensing domain). It is worth mentioning that some of the datasets have reached the scale of trillions, such as the NAIP subset of SATLAS~\cite{satlas_dataset}, or the Planet images in RapidAI4EO~\cite{rapidai4eo_dataset}, but in each case, this is over a limited region, so their size is mostly because of the very high resolution of the imagery. The (as of yet unreleased) GRAFT dataset~\cite{graft_dataset}, will approach the size of MajorTOM-Core. However, it contains only RGB Sentinel-2 data, and is missing large portions of the globe.

    Notably, many datasets~\cite{bigearthnet_dataset,seco_dataset,rapidai4eo_dataset} provide their data in a specific processed state, such as Level-2A (L2A) data for the Sentinel-2 mission, or do not state the processing level at all~\cite{satlas_dataset}. This can be limiting, since users may wish to use the original top-of-atmosphere reflectance of Level 1C. It is therefore important to provide access to the lower-level data with minimised distortions, as for example WorldStrat~\cite{worldstrat_dataset} and SSL4EO-S12~\cite{ssl4eo_dataset} do.

    Major TOM is defined as not merely a static dataset but a standard for formulating one. This manuscript also documents the release of the first large volume of open data following that standard, named MajorTOM-Core. It includes both Sentinel-2 processing levels, with a patch size of 1,068 by 1,068 pixels, and over 2.5 trillion pixels (about an order of magnitude more than the largest openly available Sentinel-2 datasets to date~\cite{ssl4eo_dataset,satlas_dataset}). It covers nearly every point on Earth captured by the Sentinel-2 mission. 

\section{Design of MajorTOM}
\label{sec:design}
    
    \subsection{Major TOM Grid}\label{ssec:grid}

        One of the foundational components of Major TOM is the grid for indexing and sampling data. Finding a uniform sampling for the surface of a sphere is not trivial. Using a uniform grid in latitude and longitude, one finds the relative spacing between points tends to zero at high latitudes, which would lead to over-sampling in higher latitude regions. Other projections, such as the \gls{utm}, work well in specific areas, with minimal distortion, but perform poorly when the longitude is outside a given range. Instead, we have defined a grid of points based on simple rules, that prioritise simplicity, generalisability, and reproducibility. The Major TOM grid is defined by a single number: the nominal grid spacing, $D$, in units of distance. We use 10 km, but other values may be used. This single parameter defines a set of rows and columns of sampling points that is the same for any data source added to Major TOM. First, the number of rows, $N_r$ is calculated as

        \begin{equation}
            N_r = \mathrm{ceil}\left(\frac{\pi R_{\bigoplus}}{D}\right)
        \end{equation}

        where $R_{\bigoplus}$ is the radius of the Earth (we use the equatorial radius of the WGS84 ellipsoid, 6378.137 km). The latitudes of these rows are then defined as evenly spaced intervals between $90^{\circ}$S and $90^{\circ}$N, such that the distance between them is almost, but not quite, $D$. We can then define this true latitudinal grid spacing, $\delta{}_{lat}$ (in degrees, not distance units), as
        
        \begin{equation}
            \delta{}_{\mathrm{lat}} = 180 / N_r
        \end{equation}

        which can be straightforwardly multiplied by a row's index to calculate it's latitude. We index the rows from 0 at the equator, with a suffix `U' (up) for north, and `D' (down) for south (up and down are used to avoid misunderstanding these indices as coordinates). So, row `317D' is at latitude $-317 \delta{}_{lat}$. For each given row, the columns' longitudes are defined independently, meaning the column indices of one row do not share longitudes with another row's. First, the row's circumference, $C_r$, of an intersecting circular plane with the Earth at the latitude of the row is given as

        \begin{equation}
            C_r = 2 \pi R_{\bigoplus} \cos\left(\frac{\pi \textrm{lat}(r)}{180}\right)
        \end{equation}

        which is used to calculate the number of columns for a row, $N^{(r)}_{c}$, in a similar way to how $N_r$ was calculated,

        \begin{equation}
            N^{(r)}_{c} = \textrm{ceil}\left(\frac{C_r}{D}\right)
        \end{equation}        

        giving us a true longitudinal spacing $\delta{}_{\mathrm{lon}}(r)$ close to our desired nominal size of $D$, 

        \begin{equation}
            \delta{}_{\mathrm{lon}}(r) = 360 / N^{(r)}_{c}
        \end{equation}

        with which, finally, we can calculate the coordinates of any grid point, say `201U, 54L' as $(201\delta{}_{lat},-54\delta{}_{\mathrm{lon}}(201U))$, where `L' and `R' are used for west and east respectively.

\begin{figure}
            \centering
            \includegraphics[width=\linewidth, trim=7cm 17cm 7cm 16cm, clip]{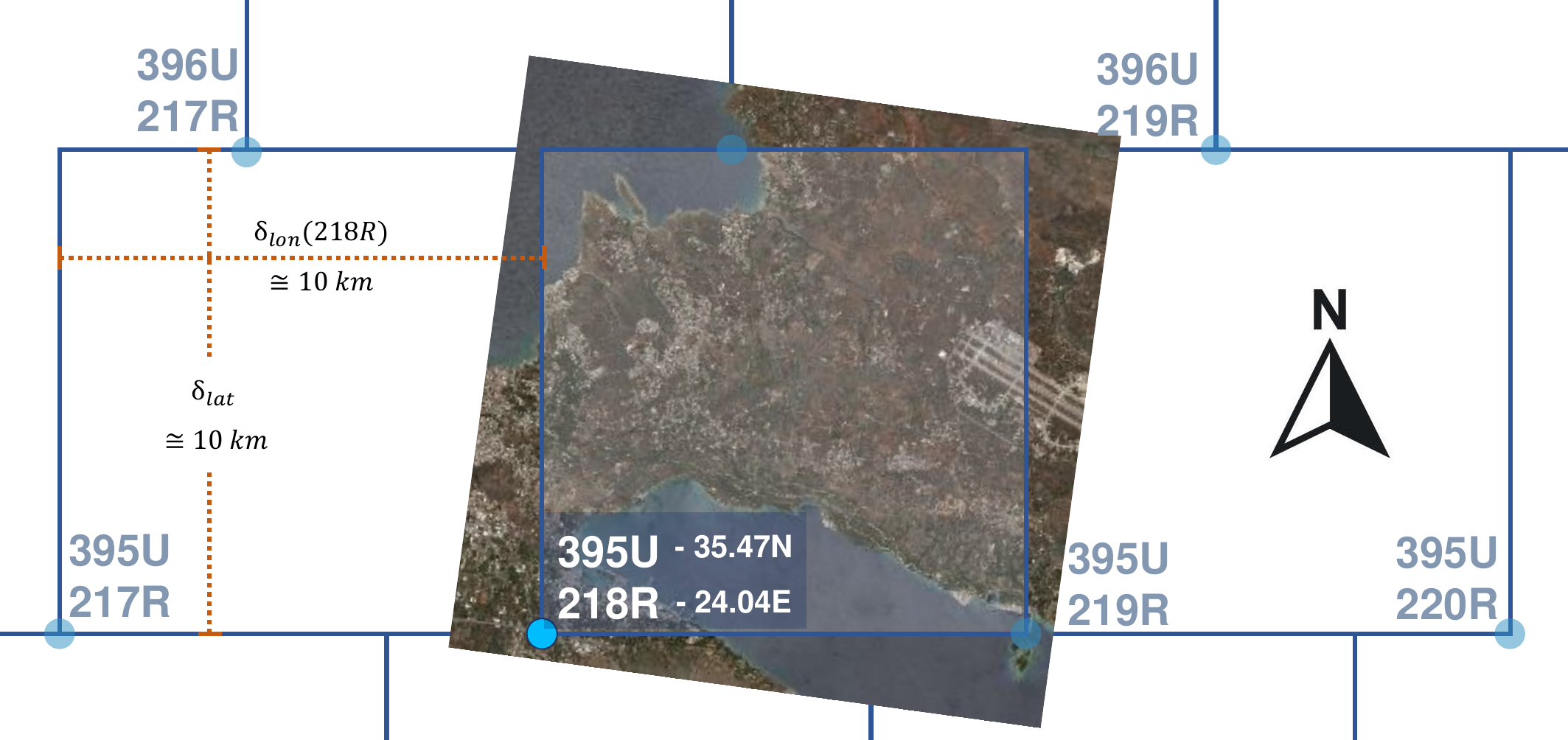}
            \caption{An example of a Major TOM grid cell over Crete, with a Sentinel-2 sample. In this case, the sample is held in a \gls{utm} projection, and overlaps slightly with other nearby grid cells.}
            \label{fig:cell-design}
        \end{figure}

        It is important to note that this grid does not define a projection, nor the exact extent of a sample. It is simply a set of points that are spaced in an approximately equidistant grid, by which data samples are indexed. Whilst these points act as anchors for data sampling in the dataset, the data sources themselves are held in their own coordinate reference systems (for many satellites, this is a \gls{utm} projection), and with their own extents (see Fig.~\ref{fig:cell-design}). The advantage of the grid is that it is computationally cheap to create, adaptable to different sizes, and has no areas of significant distortion, because of the slight differences between the nominal grid spacing and the true distances between the rows and columns.

    \subsection{Metadata}\label{ssec:guidelines}

        The specifications for Major TOM are designed to be minimally restrictive, with no dictated data formats or file structures. Rather, the goal is to offer users a way to rapidly access indexed data from different sources over the same locations, by searching and filtering the metadata in application-specific ways. Therefore, beyond data being indexed with the Major TOM grid, the only other recommendation placed on data is to provide a STAC-compliant metadata file describing the assets than can be found in each grid cell. Combining this standard with the Major TOM grid offers a practical and scalable solution for interoperable datasets. Otherwise, a metadata file with the product id, timestamp (or start and end times) and Major TOM grid cell information should be provided.
        

        This kind of metadata allows users to easily search and filter datasets in a consistent way, ensuring that multiple Major TOM datasets can be instantly combined. Furthermore, reproducing another's training and validation sets could be as simple as referencing a list of Major TOM grid points and sources.
    
\section{Major TOM Datasets}\label{sec:data}
    \subsection{MajorTOM-Core: The largest ever Sentinel-2 dataset}     

        The aim of the MajorTOM-Core dataset is to cover as wide a distribution of \gls{eo} data as possible, acting as the first contribution to the Major TOM framework. Eventually, this will include data from several satellites and data sources, however we mainly report here on the Sentinel-2 collection, as it is essentially complete. Both levels 1C and 2A were acquired using the CREODIAS service. For each point in the Major TOM grid, products covering the relevant area were found. For the globally distributed samples, the goal is to have a Sentinel-2 product covering each tile, at a random point in time.
        
        Several steps of quality control are involved in the data curation process, the most important of which is checking the cloud cover over the image. Accurate cloud masking is challenging, and the cloud mask provided with the Sentinel-2 products are prone to both false positive and false negative errors. We opted to use this mask as a rough guide for product selection, but used a state-of-the-art deep learning cloud mask~\cite{senseiv1,senseiv2} to make final predictions, which are also provided alongside the data. Samples are also checked for regions of no data. Whilst some areas with no data are inevitable close to the boundaries between different Sentinel-2 grid tiles, this was minimised wherever possible.

        When finding a suitable sample for a grid point, scenes were arranged from least to most cloudy (according to the pre-calculated Sentinel-2 cloud masks) within a randomly sampled time window of 4 months. In this order, data from each candidate scene was downloaded over the extent of the grid cell, and our deep learning-based cloud mask applied. If the total cloud cover is less than 25\%, then the data is retained. In exceptional circumstances, where 50 or more scenes are tried and no cloud-free scene can be found, then a maximum cloud coverage of 50\% is permitted. Across the dataset, there is a mean average of 4.7\% cloud cover, and a median of 0\% (more than half of samples have no cloud whatsoever). In total, 2.25 million samples have been collected, with the expectation that some of the currently unsampled points will be successfully retrieved. This amounts to a total surface area of 250 million km\textsuperscript{2} including overlapping boundaries, or 225 million km\textsuperscript{2} excluding those overlaps. This is close to 50\% of the Earth's surface, and covers almost all of the Sentinel-2 observation area. Some gaps persist, with limited coverage over the interior of Greenland, as well as some equatorial regions such as Gabon and Guyana, where cloud-free imagery is difficult to obtain.

        Learning from the past efforts documented in Sec.~\ref{sec:rel_work}, destructive preprocessing through band selection, normalisation or resampling is avoided. Patches are at a size that balance the speed of working with small files against a large field of view. The patch size of 1,068 was chosen because it is divisible by 6, meaning that bands at 10m, 20m, and 60m, are all perfectly aligned and no interpolation is needed. Pixel values remain exactly as found in the original Sentinel-2 L1C and L2A products. In addition, product diversity is maximised, with each grid cell having independent Sentinel-2 product sampling.

        Several expansions to the MajorTOM-Core dataset are currently being generated. Currently, Copernicus and Landsat missions are the main focus, due to their wide use and global coverage. Sentinel-1 data is being generated wherever possible across the grid, with products taken at the closest possible time to the ones sampled in the Sentinel-2 dataset. Based on the current processing, we estimate around 75\% of the grid cells which have Sentinel-2 data will be successfully sampled with Sentinel-1 data. 

    \subsection{MajorTOM-LUCAS-2018: Ground Control}

        Major TOM is designed not only for geolocated raster data, but also for other, heterogeneous sources of geospatial data. The LUCAS dataset~\cite{lucas_dataset} provides ground imagery and survey data at several hundred thousand locations across the European Economic Area, every 3 years. With this, a Europe-wide expansion,MajorTOM-LUCAS-2018 (otherwise known as Ground Control to Major TOM), is also being released, with paired ground-satellite imagery covering around 42k of the Major TOM grid points, and over 180k satellite patches in total. Such a dataset will provide a valuable resource for a variety of research topics, including land use land cover classification, and satellite-ground feature alignment ~\cite{graft_dataset}, demonstrating the versatility of the Major TOM framework.

    \subsection{Other Potential Uses}

        As well as data sources typically used as inputs to a model, Major TOM can also be used for the dissemination of derived products, either generated through manual labelling or model deployment. Complex, multitask problems could be easily constructed, as a greater diversity of derived products and labels are created over the same grid cells. Further, model validation results can be shared alongside published works, providing reviewers and readers the ability to quickly access and reproduce results, with a clear line between the inputs used, and the outputs generated.
        
\section{Conclusion}
\label{sec:conclusion}

    Major TOM is proposed as a more future-proof way of building \gls{eo} datasets, paving the way for a large, open, interoperable data ecosystem. Besides the MajorTOM-Core and Ground Control datasets outlined here, the Major TOM grid system will make future expansions from other sources straightforward to use interoperably. The specifications for the framework have been designed deliberately to not be overly strict, allowing future additions to flexibly adapt their specific data to the framework.

\vfill
\pagebreak


\end{document}